\documentclass{cta-author}

{}
{}
{}

\usepackage{subfig} 

\usepackage{graphics} 
\usepackage{amssymb}
\usepackage{latexsym}
\usepackage{amsmath} 
\usepackage{booktabs}

\usepackage{fancyhdr} 
\begin{document}

\title{Multi-view pose estimation with mixtures-of-parts and adaptive viewpoint selection}

\author{\au{Emre Dogan$^{1,2,3\corr}$}
\au{Gonen Eren$^3$}
\au{Christian Wolf$^{1,2}$}
\au{Eric Lombardi$^{1}$}
\au{Atilla Baskurt$^{1,2}$}
}

\address{\add{1}{Universit\'e{} de Lyon, CNRS}
\add{2}{INSA-Lyon, LIRIS, UMR CNRS 5205, F-69621, France}
\add{3}{Galatasaray University, Dept. of Comp. Eng., 36 Ciragan Cd, Istanbul 34349, Turkey}
\email{edogan@gsu.edu.tr}}

\begin{abstract}
We propose a new method for human pose estimation which leverages information from multiple views to impose a strong prior on articulated pose. The novelty of the method concerns the types of coherence modelled. Consistency is maximised over the different views through different terms modelling classical geometric information (coherence of the resulting poses) as well as  appearance information which is modelled as latent variables in the global energy function. Moreover, adequacy of each view is assessed and their contributions are adjusted accordingly. Experiments on the HumanEva and UMPM datasets show that the proposed method significantly decreases the estimation error compared to single-view results.
\end{abstract}

\maketitle

\emph{\textbf{This paper is a preprint of a paper submitted to IET Computer Vision. If accepted, the copy of record will be available at the IET Digital Library.}}

\section{Introduction}
Human pose estimation is a building block in many industrial applications such as human-computer interaction, motion capture systems, etc. 
Whereas the problem has been almost solved for easy instances, such as cooperative settings in close distance and depth data without occlusions, other realistic configurations still present a significant challenge. In particular, pose estimation from RGB input in non-cooperative settings remains a difficult problem.

Methods range from unstructured and purely discriminative approaches in simple tasks on depth data, which allow real-time performance on low-cost hardware, up to complex methods imposing strong priors on pose. The latter are dominant on the more difficult RGB data but also increasingly popular on depth. These priors are often modelled as kinematic trees (as in the proposed method) or, using inverse rendering as geometric parametric models (see section \ref{sec:relatedwork} for related works).

In this paper, we leverage the information from multiple (RGB) views to impose a strong prior on articulated pose, targetting applications such as video surveillance from multiple cameras. 
Activity recognition in this context is frequently preceded by articulated pose estimation, which --- in a non-cooperative environment such as surveillance --- can strongly depend on the optimal viewpoint. Multi-view methods can often increase robustness w.r.t. occlusions. 

In the proposed approach, kinematic trees model independent pose priors for each individual viewpoint, and additional terms favour consistency across views. The novelty of our contribution lies in the fact that consistency is not only forced geometrically on the solution, but also in the space of latent variables across views.

More precisely, a pose is modelled as mixtures of parts, each of which is assigned to a position. As in classical kinematic trees, deformation terms model relative positions of parts w.r.t. neighbours in the tree. In the lines of \cite{262_yangramanan2013}, the deformations and the appearance terms depend on latent variables which switch between mixture components. This creates a powerful and expressive model with low-variance mixture components which are able to model precise relationships between appearance and deformations. Intuitively, and as an example, we could imagine relative positions of elbow and forearm to depend on a latent variable, which itself depends on the appearance of the elbow. It is easy to see that a stretched elbow requires a different relative position than a bent elbow.

In the proposed multi-view setting, positions, as well as latent variables, are modelled for each individual view. A global energy term favours a consistent pose over the complete set of views, including consistency of the latent part type variables which select mixture components. Here the premise is that appearance may certainly differ between viewpoints, but that a given pose is translated into a subset of consistent latent mixture components which can be learned. 

\begin{figure*}[t]
    \centering{
    \includegraphics[width=0.9\textwidth]{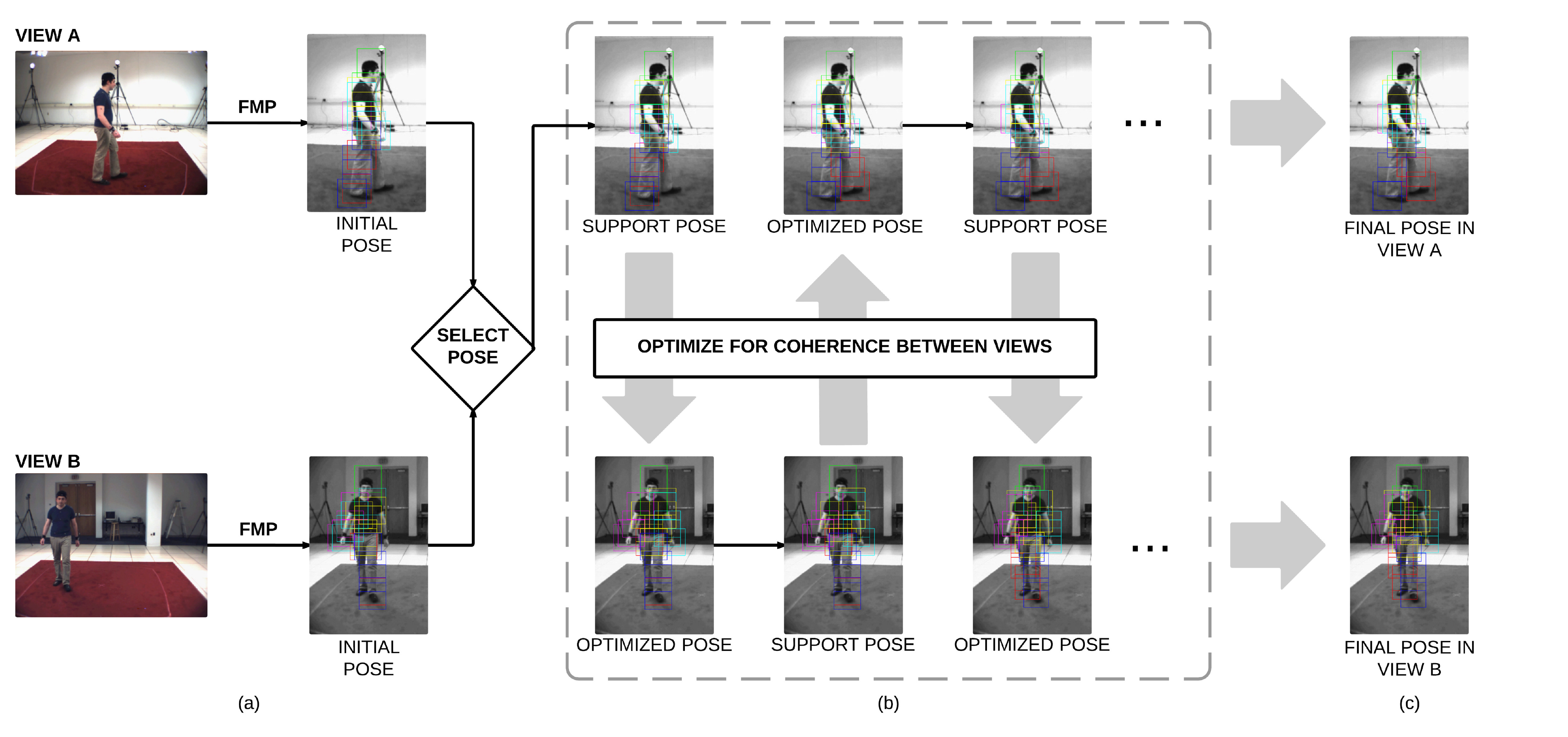}}
    \caption{Overview of the model: a) Initial pose estimation running the single-view model on each view separately. The pose with the highest confidence score is selected as the support pose. b) Joint estimation loop with geometrical and appearance constraints provided by support pose. The newly obtained pose becomes the support pose at the end of each iteration and provides constraints for the other view. c) After convergence, the last two poses are returned as the final results. Best viewed in colour.}
    \label{fig_overview}
\end{figure*}

An overview of the proposed method can be seen in Fig. \ref{fig_overview} which depicts the iterative nature of the multi-view pose estimation process. Basically, one of the single-view estimations is selected as the support pose and provides additional information to the other view. On each iteration, \emph{support} and \emph{target} poses are swapped so that both predictions improve over iterations. The optimisation loop continues until convergence, where a final pose is produced for each view.

As a summary, our main contributions are the following:
\begin{itemize}
\item We propose a global graphical model over multiple views including multiple constraints: geometrical constraints and constraints over the latent appearance space.
\item We propose an iterative optimization routine.
\item We introduce an adaptive viewpoint selection method, which removes viewpoints if the expected error is too high.
\end{itemize}

%
%
\section{Related Work}
\label{sec:relatedwork}

\noindent
Human pose estimation from RGB images has received increasing attention, we therefore restrict this section by excluding techniques that exploit depth images and focus on part-based models, generative and discriminative probabilistic models, tracking methods and deep neural networks.

\textit{Pictorial structures (PS)} --- are a dominant family of models. Based on the original idea in \cite{250_pictorialstructures}, they model an object as a combination of parts related to a star-shaped or tree-shaped structure and deformation costs. The problem is formulated as an energy function with terms for appearance similarity plus deformation terms between parts \cite{319_sappJordanTaskar, 315_dantone_pami2014}. Although efficient for inference, tree-structured models are known to suffer from the double-counting problem, especially for limb parts. To address this issue loopy constraints are commonly used, but they require diverse approximate inference strategies \cite{331_NIPS2007_3271,323_Zhang_2015_ICCV,268_cherian_2014_CVPR}. The relationship between the non-adjacent body parts is discussed in \cite{257_pishchulin_2013_CVPR} where a fully-connected model is proposed. Due to mid-level representations conveyed by poselets, unary and binary terms are updated during test time and the model is reduced to a classical PS, which can be solved directly. \cite{269_kiefel2014eccv} proposes a PS-inspired model, with binary random variables for each part, they model the presence and absence for every position, scale and orientation. However, this results in a high number of variables which forces them to approximate inference.

\textit{Flexible Mixtures of Parts (FMP)} --- have been introduced by \cite{262_yangramanan2013}, to tackle the limited expressiveness of tree-shaped models. Instead of orientations of parts, they proposed mixtures, which are obtained by clustering appearance information. A detailed review of the method can be found at section \ref{sec:single_view_pe}. Among extensions, \cite{254_Eichner2013} proposed appearance sharing between independent samples to cluster similar background and foregrounds. \cite{267_wang_2014_CVPR} presented a method to estimate 3D pose from a single image where they use FMP and camera parameter estimation, in conjunction with anthropomorphic constraints. \cite{292_cho_spl2015} proposed an improvement by aggregation of multiple hypotheses from the same view using kernel density approximation. 

\textit{Multi-view settings} --- allow various methods to emerge, particularly for 3D pose estimation. \cite{284_sigal2011} proposed a generalised version of PS, which also exploits temporal constraints. They employ graphs spanning multiple frames and inference is performed with non-parametric belief propagation. Among 3D extensions of PS, the burden of the infeasible 3D search space is handled by reducing it with discretisation \cite{279_burenius2013}, supervoxels \cite{301_schick2015}, triangulation of corresponding body parts from different viewpoints \cite{314_belagiannis2015} or by using voxel-based annealing particle filtering \cite{302_cantonferrer2009}. Recently, \cite{300_zuffi2015} proposed a strategy similar to 3D-PS, but with a realistic body model, and inference is carried out with particle-based max-product belief propagation. Inferring 3D pose from multiple 2D poses is also common, with various underlying strategies such as hierarchical shape matching \cite{287_hoffmanAndGavrila}, random forests \cite{280_kazemi2013} and optical flow \cite{285_puwein2014}. \cite{275_aminbmvc2013} introduced a scheme where PS is employed to estimate 2D poses, then incorporates them to obtain a 3D pose with geometrical constraints, colour and shape cues. Although being somewhat analogous to our proposition, \cite{275_aminbmvc2013} does not consider cases where some viewpoints are more beneficial than others, which we leverage with adaptive viewpoint selection.

\textit{Temporal strategies} --- are commonly used for pose estimation and articulated tracking from videos. Using spatiotemporal links between the individual parts of consecutive frames seems promising, but intractability issues arise. To this end, \cite{268_cherian_2014_CVPR} opt for approximation with distance transform \cite{felzenszwalb2012distance}. \cite{323_Zhang_2015_ICCV} reduce the graph by combining symmetrical parts of human body and generating part-based tracklets for temporal consistency. \cite{325_Nie_2015_CVPR} uses a spatiotemporal And/Or Graph to represent poses where only temporal links exist between parts. Recently \cite{326_ParkRamanan_2015_CVPR} proposed synthesising hypotheses by applying geometrical transformations to initially annotated pose and match next frame with the nearest neighbour search.

\textit{Discriminative approaches} --- learn a direct mapping from feature space to pose, often by avoiding any explicit body models (although models can be integrated). Silhouettes \cite{329_agarwal-triggs_pami06} and edges \cite{309_bo2008} are frequently used as image features in conjunction with learning strategies for probabilistic mapping, such as regression \cite{329_agarwal-triggs_pami06}, Gaussian Processes \cite{294_urtasun2008} and mixtures of Bayesian Experts \cite{331_NIPS2007_3271}. Previous work shows that these approaches are usually computationally efficient and perform well in controlled environments, but they are highly dependent on the training data and therefore tend to generalise poorly in unconstrained settings.

\textit{Deep neural networks} --- have received remarkable attention recently which inevitably affected the pose estimation challenge. \cite{253_ouyang_2014_CVPR} address the problem by first obtaining 2D pose candidates with PS, then utilising a deep model the determine the final pose. \cite{324_Fan_2015_CVPR} feeds both local patches and their holistic views into the Convolutional Neural Networks (CNN), while \cite{321_yannlecunNIPS2014} proposes a new architecture where deep CNNs are used in conjunction with Markov Random Fields. \cite{263_toshev_2014_CVPR} on the other hand, follow a more direct approach and employ a cascade of Deep Neural Network regressors to handle the pose estimation task. \cite{ChenYuille2014} uses a kinematic tree, where the same deep network learns unary terms as well as data dependent binary terms. Similar to our adaptive viewpoint selection scheme, predicting the estimation error during test time is explored by \cite{466_Carreira_2016_CVPR}; however they employ a CNN to learn iterative corrections that converge towards the final estimation. Part mixtures is adopted in \cite{471_Yang_2016_CVPR, 465_Chu_2016_CVPR}, where message passing is implemented as additional layers. State-of-the-art results on single-view benchmarks are achieved by \cite{479_newell2016_ECCV}, where multiple hourglass layer modules are stacked end-to-end.

Our method is based on FMP, which allows imposing a strong prior on pose, generalising it to multiple views. Compared to existing multi-view approaches, our method is not restricted to geometric coherence terms. We enforce coherence also in the space of latent variables (the mixture component selectors), which further improves performance. Additionally, we leverage the consistency between views by predicting the fitness of each view during test time.
\section{Single view pose estimation} 
\label{sec:single_view_pe}

\noindent
In the lines of \cite{262_yangramanan2013}, an articulated pose in a single 2D image is modelled as a flexible mixture of parts (FMP). Related to deformable part models introduced by \cite{147_Felzenszwalb_pami2010}, part-based models for articulated pose estimation are classically tree structured. Similarly, our model is a kinematic tree on which a global energy is defined including data attached unary terms and pairwise terms acting as a prior on body pose. The underlying graph is written as $G=(V,E)$, where vertices are parts and edges are defined on adjacency between parts.

Let $p_i = (x,y)$ be the pixel coordinates for part $i \in \{1,\ldots,K\}$ in image $I$. The values of $p_i$ are the desired result values over which we would like to optimise. Additional latent variables $t_i \in \{1,\ldots,T\}$ model a type of this part, which allows to model terms in the energy function for given types, effectively creating a powerful mixture model. In practice, the part types are set to clusters of appearance features during training time.
In the single-view version, the energy function corresponds to the one given in \cite{262_yangramanan2013}. Defined over a full pose $p=\{p_i\}$, input image $I$ and latent variables $t=\{t_i\}$, it is given as follows:
\begin{equation} \label{eq:yr_fullModel}
\begin{array}{ll}
S(I,p,t) &= \sum_{i \in V}w_{i}^{t_i} \phi(I, p_i) + \sum_{ij \in E}w_{ij}^{t_i,t_j} \psi(p_i - p_j) \\[5pt]
         & +\sum_{i \in V}b_{i}^{t_i} + \sum_{ij \in E}b_{ij}^{t_i,t_j}
\end{array}
\end{equation}
The expression in the first sum corresponds to data attached terms, where $\phi(I,p_i)$ are appearance features extracted at $p_i$ (HoG, see section \ref{sec:experiments}). Note that the corresponding trained parameters $w_{i}^{t_i}$ depend on the latent part type $t_i$. 

The pairwise terms in the next expression model the prior over body pose using a classical second degree deformation $\psi(p_i - p_j) = [  dx \ dx^2 \ dy \ dy^2  ]^T$ where $dx = x_i{-}x_j$ and $dy = y_i{-}y_j$. They control the relative positions of the parts and act as a ``switching'' spring model, the switching controlled by the latent part types $t_i$.

The last two sums define a prior over part types including unary part type biases $b_{i}^{t_i}$ and pairwise part type terms $b_{ij}^{t_i,t_j}$. 

Although scale information is not specified in the equations, a pyramid is used in the implementation to address the various sizes of the subjects in the image. Inference in this model (and in our generalisation to multi-view problems) will be addressed in section \ref{sec:inference}.

\begin{figure}[tb]
\centering{
\subfloat[]{\label{subfigure:p15t4} \includegraphics[width=0.3\columnwidth]{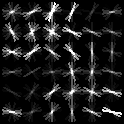}}
\subfloat[]{\label{subfigure:p15t1} \includegraphics[width=0.3\columnwidth]{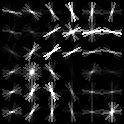}}
\subfloat[]{\label{subfigure:p15t2} \includegraphics[width=0.3\columnwidth]{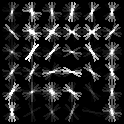}}
}
\caption{Illustration of the learned multi-view consistency term over latent appearance (part types). The picture shows HoG filters of the same part (the shoulder). Filter pair (a -- b) has been learned to be highly compatible (eventually across different viewpoints), whereas compatibility of pair (a -- c) has been learned to be low, according to training data.}
\label{fig:hogFilters}
\end{figure}

\section{Multi-view pose estimation}
\label{sec:multiview}

\noindent
We generalise the single-view model to multiple views and show that geometrical consistency constraints can be leveraged to improve estimation quality. Without loss of generality let us fix the number of views to two and consider a setup with calibrated cameras. In this case, a global energy function models the pose quality over a pair views $A$ and $B$, taking as input images $I^A$ and $I^B$ and estimating pose variables $p^A$ and $p^B$ while additionally optimising over latent part types $t^A$ and $t^B$:
\begin{equation} \label{eq:our_model_full}
\begin{array}{ll}
S(I^A, I^B, p^A, p^B, t^A, t^B) &= \\[5pt]
    S(I^A, p^A, t^A) &+ \ S(I^B, p^B, t^B) \\[5pt]
    + \alpha\sum_{i \in V} \boldsymbol{a_i} \xi(p_{i}^{A}, p_{i}^{B}) &+ \ \beta\sum_{i \in V} \boldsymbol{a_i} \lambda(t_{i}^{A}, t_{i}^{B}) \\
\end{array}
\end{equation}
Here, $S$ is the single pose energy from equation \ref{eq:yr_fullModel}. The two additional terms $\xi$ and $\lambda$ ensure consistency of the pose over the two views, $\alpha$ and $\beta$ are the hyperparameters that controls the amount of influence of these terms, and $a_i$ are binary variables activating or deactivating consistency. All terms and symbols are described in the corresponding sections below.

\subsection{Geometric constraints}
\label{subsec:geometric}
Assuming temporal synchronisation, images $I^A$ and $I^B$ show the same articulated pose from two different viewpoints, which can be exploited. In particular, given calibrated cameras, points in a given view correspond to epipolar lines in the second view. The geometric term $\xi$ leverages this as follows:
\begin{equation}
\xi(p_{i}^{A}, p_{i}^{B}) = -d(p_{i}^{A}, e(A,p_{i}^{B})) - d(p_{i}^{B}, e(B,p_{i}^{A}))
\end{equation}
where $e(A,p_i^{B})$ is the epipolar line in view $A$ of point $p_i$ in view $B$ and $d$ is the Euclidean squared distance between a point and a line. Thus, geometric constraints translate as additional energy to particular locations for both views in the global energy function.

\subsection{Appearance constraints}
\label{subsec:appearance}
The geometric constraints above are imposed on the solution (positions $p_i$). The term $\lambda$ adds additional constraints on the latent part type variables $t_i$, which further pushes the result to consistent solutions. Recall that the latent variables are clusters in feature space, i.e. they are related to types of appearance. Appearances might, of course, be different over views as a result of the deformation due to viewpoint changes. However, some changes in appearances will likely be due to the viewpoint change, whereas others will not. Intuitively, we can give the example of an open hand in view $A$, which will certainly have a different appearance in view $B$; however, the image will not likely be the one of a closed hand.

We model these constraints in a non-parametric way as a discrete distribution learned from training data, i.e. $\lambda(t_{i}^{A}, t_{i}^{B}) = p(t_{i}^{A}, t_{i}^{B})$ (see section \ref{sec:training}). {Figure {\ref{fig:hogFilters} illustrates this term using three filter examples shown for the learned model of part \emph{right shoulder}. The $\lambda$ term is high between (\ref{subfigure:p15t4}) and (\ref{subfigure:p15t1}), but low between (\ref{subfigure:p15t4}) and (\ref{subfigure:p15t2}). Intuitively, (\ref{subfigure:p15t4}) and (\ref{subfigure:p15t1}) look like the same 3D object seen from different angles, whereas (\ref{subfigure:p15t4}) and (\ref{subfigure:p15t2}) do not.

\subsection{Adaptive viewpoint selection}
\label{subsec:wfv}
Geometric and appearance constraints rely on the accuracy of the initial single-view pose estimates. 
In certain cases, the multi-view scheme can  propagate poorly estimated part positions over views, eventually deteriorating the multi-view result. To solve this problem, we would like to estimate beforehand, whether an additional view can contribute, i.e. increase performance, or whether it will deteriorate good estimations from a better view.

We propose an adaptive viewpoint selection mechanism and introduce a binary indicator vector (over parts) that switches on and off geometric and appearance constraints for each part during inference. If an indicator is switched off for a part, then the support pose does not have an effect on the optimised pose for this part. The binary indicator vector $\boldsymbol{a}$ is given as follows:
\begin{equation}
\boldsymbol{a}_i = \left \{
\begin{array}{cc} 
0 & \textrm{if }\sigma_i(p^A,\theta) > \tau_i \ \textrm{or }\sigma_i(p^B,\theta) > \tau_i \\
1 & \textrm{else}\\
\end{array}
\right .
\end{equation}
where $\tau_i$ is a threshold obtained from median part errors on the training set and $\sigma_i(p^A,\theta)$ is a function with parameters $\theta$ that estimates the expected error committed by the single-view method for part $i$, given an initial estimate of the full pose $p^A$. 

$\sigma$ is a mapping learned as a deep CNN taking image tiles cropped around the initial (single-view) detection $p^A$ as input. Training the network requires to minimise a loss over part estimation errors, i.e. an error over errors, as follows:
\begin{equation}
\displaystyle{
\min_\theta || \ \sigma(p^A,\theta) - {\mathbf e} \ ||_2
}
\end{equation}
where ${\mathbf e}$ is the vector of ground truth errors obtained for the different parts by the single-view method, and $||\cdot||_2$ is the $L_2$ norm which is here taken over a vector holding estimations for individual parts. $\theta$ are the parameters of the deep network.

We argue that such a network is suitable to anticipate whether an individual part is useful for multi-view scheme, by implicitly learning multi-level features from an image tile. For example, self-occluded parts or other poor conditions would most likely to be associated with high error rates, whereas unobstructed views would yield low errors. Thresholding the output of the network, namely the error estimations ${\sigma}_i$, can provide the decision whether the support view has an influence for part $i$ or not.

\section{Training}
\label{sec:training}


\paragraph{Single-view parameters} 
Appearance coefficients $w_{i}^{t_i}$, deformation coefficients $w_{ij}^{t_i,t_j}$ and part type prior coefficients $b_{i}^{t_i}$ and $b_{ij}^{t_i,t_j}$ are learned as in \cite{262_yangramanan2013}: we proceed by supervised training with positive and negative samples, where the optimisation of the objective function is formulated as a structural SVM. Part type coefficients are learned w.r.t. their relative positions to their parents by clustering. This mixture of parts approach ensures the diversity of appearances of part types where their appearance is associated with their placement with reference to their parents; for example a left-oriented hand is usually seen on the left side of an elbow, while a upward facing hand is likely to occur above an elbow.  

\paragraph{Consistency parameters} The discrete distribution $\lambda(t_{i}^{A}, t_{i}^{B})= p(t_{i}^{A}, t_{i}^{B})$ related to the appearance constraints between views is learned from training data as co-occurrences of part types between the viewpoint combinations. We propose a weakly-supervised training algorithm which supposes annotations of the pose (positions $p_i$) only, and which does not require ground truth of part types $t_i$. In particular, the single-view problem is solved on the images of two different viewpoints and the resulting poses are checked against the ground truth poses. If the error is small enough, the inferred latent variables $t_i$ are used for learning. The distribution $p(t_{i}^{A}, t_{i}^{B})$ is thus estimated by histogramming eligible values for $t_i^{A}$ and $t_i^{B}$. Fig \ref{fig:hogFilters} shows an example of learned filters and their compatibility.

The hyper-parameters $\alpha$ and $\beta$ weighting the importance of the consistency prior are learned through cross-validation over a hold-out set (see section \ref{sec:experiments}).

\paragraph{Viewpoint selection parameters} As seen in Section \ref{subsec:wfv}, $\sigma$ is a mapping that estimates error of a single-view pose estimation, given an image tile cropped around the bounding box. To determine $\sigma$, we use regression of the expected error and train a deep CNN. We use a VGG-16 network \cite{500_SimonyanVGG2014_ICLR} pre-trained on \emph{ImageNet} and remove all the top fully connected layers and replace them with a single small hidden layer for regression. We finetune the last convolutional block of VGG and learn the weights of the newly added fully connected layers with augmented data (see section \ref{sec:experiments} for further details).

\begin{figure*}[tb]
    \centering
    \includegraphics[width=0.9\textwidth]{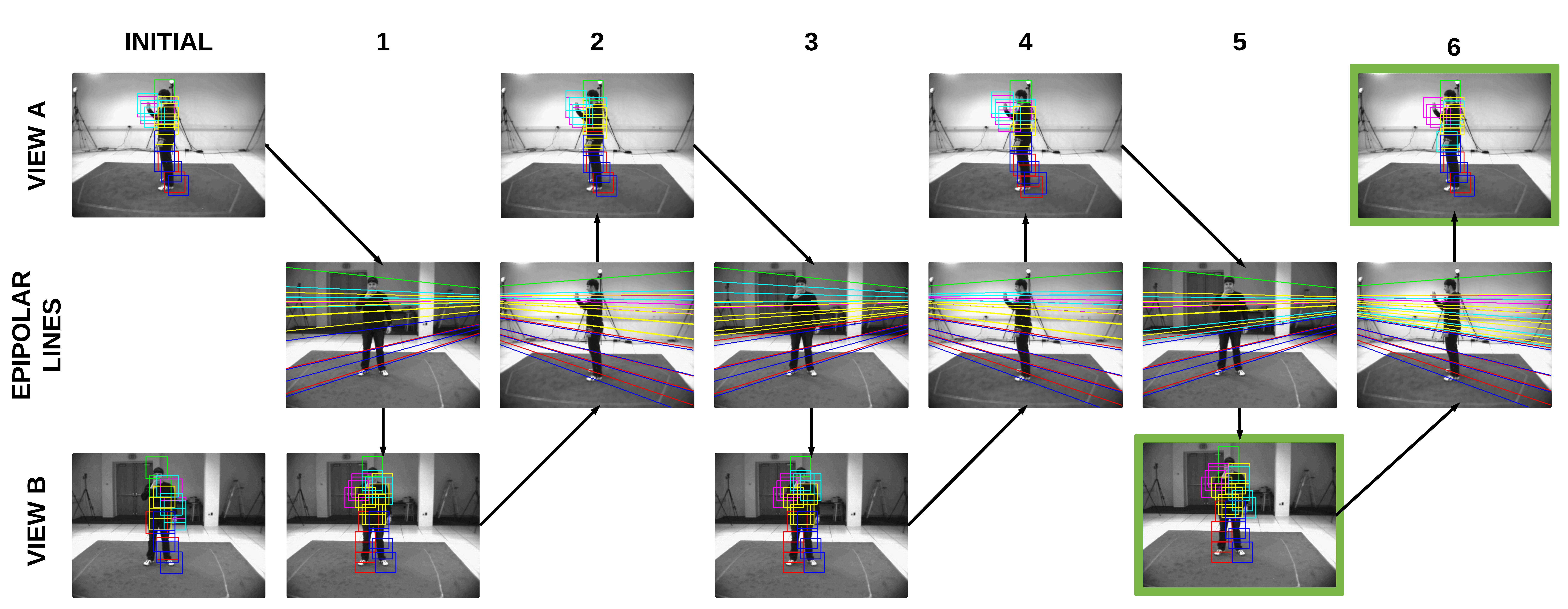}
    \caption{Illustration of the iterative optimisation process. The first and last rows are two respective viewpoints, the middle row shows epipolar lines overlaid over the respective viewpoint. Diagonal arrows show the pose that the epipolar lines are based on. Each column is an iteration and vertical arrows shows the resulting pose and epipolar lines used in joint estimation. Final poses are marked with green borders. Best viewed in colour.}
    \label{fig_alliterations}
\end{figure*}

\begin{table}[tb]
\processtable{HumanEva -- PCP3D scores (\%) of our model trained on subject 1, evaluated on subject 1 and all subjects combined, with PCP threshold $0.5$. Performance of  is compared to Flexible Mixture of Parts (FMP) \cite{262_yangramanan2013} method.\label{tab:humaneva-yrour}}
{
\vspace{2pt}
\centering
\begin{tabular}{cccccc}
\hline \\[-5pt]
Subject & Sequence & FMP \cite{262_yangramanan2013} & 
\multicolumn{3}{c}{------------ Ours ------------} \\
 & & & {Geom.} & {+App.} & {+Adap.}\\
\hline \\[-5pt]
S1   & Box       & 77.34 & 82.70 & 83.87 & 85.31  \\
All  & Box       & 67.14 & 69.45 & 70.23 & 71.57  \\
S1   & Gestures  & 78.91 & 84.27 & 84.08 & 88.14  \\
All  & Gestures  & 74.68 & 77.38 & 78.81 & 80.34  \\
S1   & Jog       & 84.91 & 86.75 & 86.70 & 86.86  \\
All  & Jog       & 77.52 & 80.16 & 79.84 & 80.97  \\
S1   & Walking   & 84.65 & 86.71 & 86.50 & 87.68  \\
All  & Walking   & 78.49 & 81.69 & 81.96 & 83.17  \\
\hline \\[-5pt]
S1          & Overall   & 82.02 & 85.43 & 85.49 & 87.24 \\
All  & Overall   & 74.86 & 77.62 & 78.11 & 79.40 \\
\hline
\end{tabular}}{}
\end{table}

\begin{table}[tb]
\processtable{UMPM -- PCP3D scores (\%) on all sequences with PCP threshold $0.5$, compared to Flexible Mixture of Parts (FMP) \cite{262_yangramanan2013} method.\label{tab:umpm-yrour-All}}
{
\vspace{2pt}
\centering
\begin{tabular}{ccccc}
\hline \\[-5pt]
Sequence & FMP \cite{262_yangramanan2013} & 
\multicolumn{3}{c}{------------ Ours ------------} \\
 & & {Geom.} & {+App.}& {+Adap.}\\
\hline \\[-5pt]
Chair    & 74.72 & 78.09 & 77.54 & 79.94 \\
Grab     & 74.23 & 76.25 & 77.18 & 81.92 \\
Orthosyn & 72.47 & 74.65 & 75.22 & 76.48 \\
Table    & 70.30 & 73.49 & 74.18 & 77.86 \\
Triangle & 73.69 & 77.26 & 77.81 & 83.81 \\
\hline \\[-5pt]
Overall  & 73.07 & 75.91 & 76.37 & 80.04 \\
\hline
\end{tabular}}{}
\end{table}

\begin{table*}[tb]
\processtable{PCP3D scores (\%) for all limb parts with PCP threshold $0.5$, compared to FMP\cite{262_yangramanan2013} on HumanEva and UMPM datasets.\textit{(U-L: upper left, U-R: upper right, L-L: lower left, L-R: lower right)}\label{tab:partBasedPCP3Dscores}}
{
\centering
\begin{tabular}{ccccccccc}
\hline \\[-5pt]
Configuration & U-R Arm & U-L Arm & L-R Arm & L-L Arm & U-R Leg & U-L Leg & L-R Leg & L-L Leg \\
\hline \\[-5pt]
FMP\cite{262_yangramanan2013} on HumanEva   & 88.4 & 83.4 & 51.8 & 61.4 & 100 & 100 & 73.6 & 67.9 \\ 
Ours on HumanEva                            & 94.5 & 88.3 & 76.1 & 71.5 & 100 & 100 & 82.7 & 73.6 \\ 
\hline \\[-5pt]
FMP\cite{262_yangramanan2013} on UMPM   & 50.6 & 50.7 & 31.3 & 28.6 & 99.4 & 98.6 & 78.4 & 64.6 \\ 
Ours on UMPM                            & 69.8 & 63.6 & 45.2 & 35.6 & 99.6 & 99.5 & 84.4 & 75.1 \\ 
\hline
\end{tabular}}{}
\end{table*}

\section{Inference}
\label{sec:inference}
\noindent
Inference of the optimal pose pair requires maximising equation (\ref{eq:our_model_full}) over both poses $p_i^{A}$ and $p_i^{B}$ and over the full set of latent variables $t_i^{A}$ and $t_i^{B}$. Tractability depends on the structure of the graph, and on the clique functionals. Whereas the graph $G=(V,E)$ for the single-view problem (the graph underlying equation \ref{eq:yr_fullModel}) is a tree, the graph of the multi-view problem contains cycles. This can be seen easily, as it is constructed as a union of two identical trees with additional edges between corresponding nodes, which are due to the consistency terms. Compared to the single-view problem, maximisation cannot be carried out exactly and efficiently with dynamic programming.

Several strategies are possible to maximise equation (\ref{eq:our_model_full}): approximative message passing (loopy belief propagation) is applicable for instance, which jointly optimises the full set of variables in an approximative way, starting from an initialisation. We instead chose an iterative scheme which calculates the \emph{exact} solution for a subset of variables keeping the other variables fixed, and then alternates. In particular, as shown in figure \ref{fig_overview}, we optimise for a given view while keeping the variables of the other view (the ``support view'') fixed. Removing an entire view from the optimisation space ensures that the graph over the remaining variables is restricted to a tree, which allows solving the sub-problem efficiently using dynamic programming.

Let us write $kids(i)$ for the child nodes of part $i$. The score of a part location $p_i$ for a given part type $t_i$ is computed as follows:
\begin{equation} \label{eq:our_score}
\begin{split}
\textnormal{score}_i(t_{i}^{A},t_{i}^{B}, p_{i}^{A}, p_{i}^{B}) &= b_{i}^{t_{i}^{A}} + w_{i}^{t_i}\cdot\phi(I^A, p_{i}^{A}) \\
 &+ b_{i}^{t_{i}^{B}} +  w_{i}^{t_i}\cdot\phi(I^B, p_{i}^{B}) \\
 &+ \boldsymbol{a_i} (\alpha\xi(p_{i}^{A}, p_{i}^{B}) + \beta\lambda(t_{i}^{A}, t_{i}^{B})) \\
 &+ \sum_{k \in kids(i)}m_k(t_{i}^{A}, t_{i}^{B}, p_{i}^{A}, p_{i}^{B})
\end{split}
\end{equation}
with the message that part $i$ passes to its parent $j$ is defined as:
\begin{equation} \label{eq:our_message}
\begin{split}
m_i(t_{j}^{A}, t_{j}^{B}, p_{j}^{A}, p_{j}^{B}) &= 
\max_{t_{i}^{A}, t_{i}^{B}} \Big[ b_{ij}^{t_{i}^{A},t_{j}^{A}} +  b_{ij}^{t_{i}^{B},t_{j}^{B}} \\
&+ \max_{p_{i}^{A}, p_{i}^{B}}\textnormal{score}_i(t_{i}^{A},t_{i}^{B}, p_{i}^{A}, p_{i}^{B}) \\
&+ w_{ij}^{t_{i}^{A}, t_{j}^{A}}\cdot\psi(p_{i}^{A}, p_{j}^{A}) 
+ w_{ij}^{t_{i}^{B}, t_{j}^{B}}\cdot\psi(p_{i}^{B}, p_{j}^{B}) \Big]
\end{split}
\end{equation}
As mentioned, one of the two sets $A$ and $B$ is kept constant at each time, which simplifies the equations (\ref{eq:our_score}, \ref{eq:our_message}) to a single-view form, similar to \cite{262_yangramanan2013}.  Messages from all children of part $i$ are collected and summed with the bias term and filter response, resulting in the score for that pixel position and mixture pair. As classically done in deformable parts based models, the optimisation can be carried out with dynamic programming and the inner maximisation in equation (\ref{eq:our_message}) with min-convolutions (a distance transform, see \cite{felzenszwalb2012distance}).  

The algorithm is initialised by solving the single-view problem independently for each viewpoint. The pose with the lowest estimated error (see section \ref{subsec:wfv}) is chosen as initial support pose, the pose of the other viewpoint being optimised in the first iteration. The iterative process is repeated on until convergence or a maximum number of iterations is reached. Optimising each sub-problem is classical, where the message passing scheme iterates from the leaf nodes to the root node. After thresholding to eliminate weak candidates and non-maximum suppression to discard similar ones, backtracking obtains the final pose. 

\section{Experiments}
\label{sec:experiments}

\noindent
We evaluated our work on two datasets, \emph{HumanEva I} \cite{313_sigal2010humaneva} and \emph{Utrecht Multi-Person Motion (UMPM)} \cite{UMPM}. Both datasets have been shot using several calibrated cameras. Ground truth joint locations were recorded with a motion capture system, with 20 and 15 joints, respectively. 

For HumanEva set we only use three cameras (C1, C2 and C3), which are placed with 90 degrees offset. Three subjects (S1, S2 and S3) perform following activities: walking, boxing, jogging, gestures and throw-catch. There are three takes for each sequence, used for training, validation and test. Since the creators of HumanEva favour online evaluation, original test set does not contain ground truth joint positions. Following \cite{275_aminbmvc2013}, we divided the original training set into training and validation sets and used the original validation set for testing. All hyper-parameters have been optimised over our validation set. 

For UMPM set, all available cameras (F, L, R and S) were used. We considered all available sequences with one subject, which includes object interactions such as sitting on a chair, picking up a small object, leaning and lying on a table. The training, validation and test partitions were divided using 60\%, 20\% and 20\% of the all available data, respectively. The HumanEva test set consists of 4493 images per camera, while UMPM test set has 6074 images per camera. The number of distinct images used in the tests sums up to 13479 and 24296, respectively.

Since our model is trained with 26 parts, we used a linear function to convert our box centres to the 20 joint locations for HumanEva and 15 joints locations for UMPM.

The data attached terms $\phi(.,.)$ in this work were based on HoG features from \cite{169_HOG_dalaltriggs}. Other features are possible, in particular learned deep feature extractors as in \cite{ChenYuille2014} or \cite{NeverovaArxiv2016handpose,FourureNeurocomp2017}. This does not change the setup, and can be performed with finetuning of a pre-trained model for this case, where the amount of training data is relatively low.

We evaluate our multi-view approach against the single-view method given in \cite{262_yangramanan2013}. We use two poses as input and evaluate on one of these two poses, varying over multiple configurations.

Parameters of the single-view model (Eq. \ref{eq:yr_fullModel}) are learned on all activities of S1 for HumanEva. We took 100 frames with equal time intervals for every activity from three cameras for training, which sums up to 1500 images. The remainder of the data was set as the validation set. For UMPM, nearly 400 consecutive frames for each sequence were used as positive samples. As for the negative samples, background images from corresponding datasets were used in addition to the \emph{INRIA Person Database} \cite{169_HOG_dalaltriggs}.  Hyper-parameters $\alpha$ and $\beta$ of equation (\ref{eq:our_model_full}) were learned on validation sets. 

\begin{figure*}[tb]
\centering{
\subfloat[HumanEva, S1, all sequences]{\label{fig:pcp3dBar_he}\includegraphics[width=.475\textwidth]{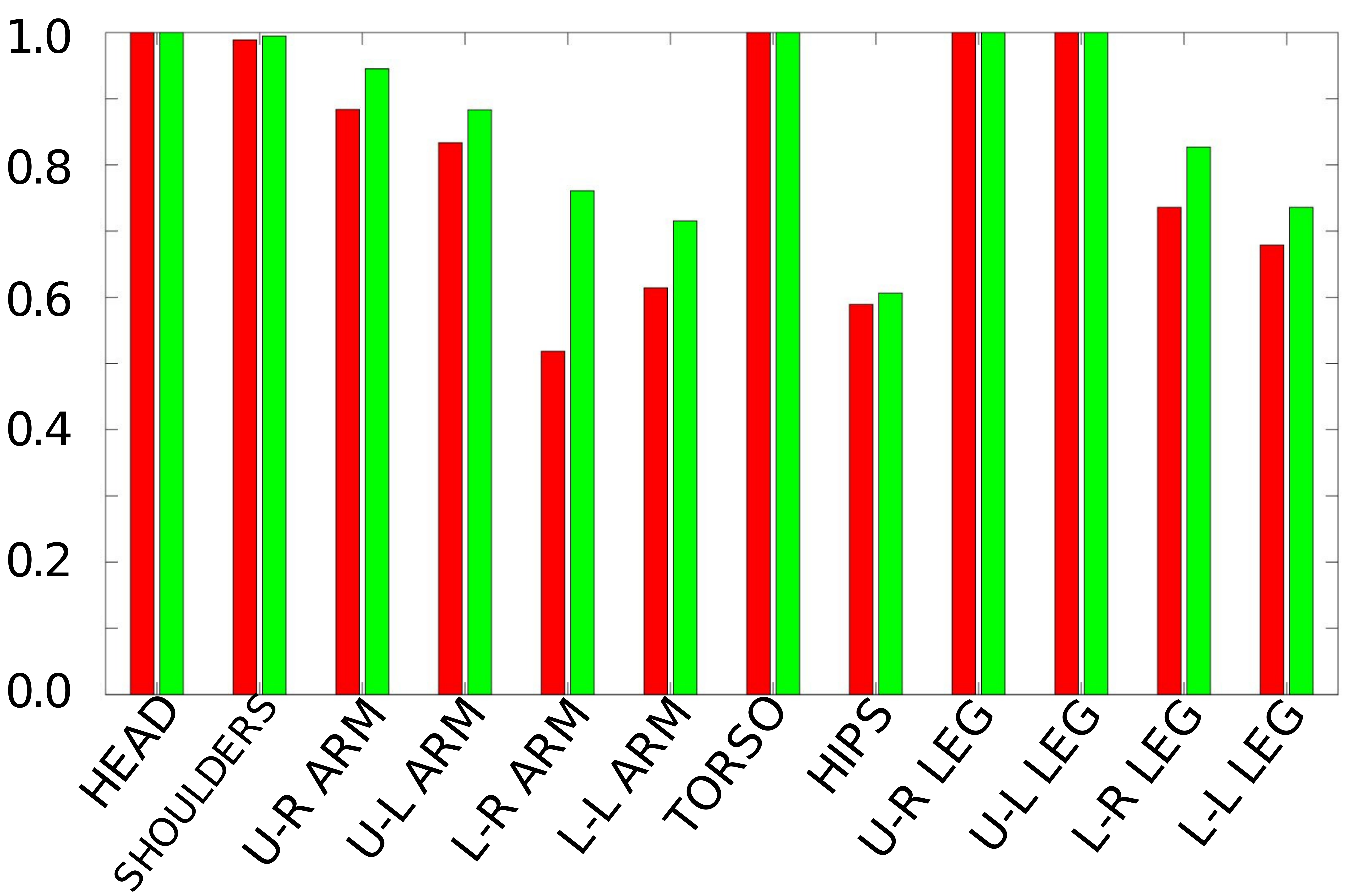}}
\subfloat[UMPM, all sequences]{\label{fig:pcp3dBar_umpm}\includegraphics[width=.475\textwidth]{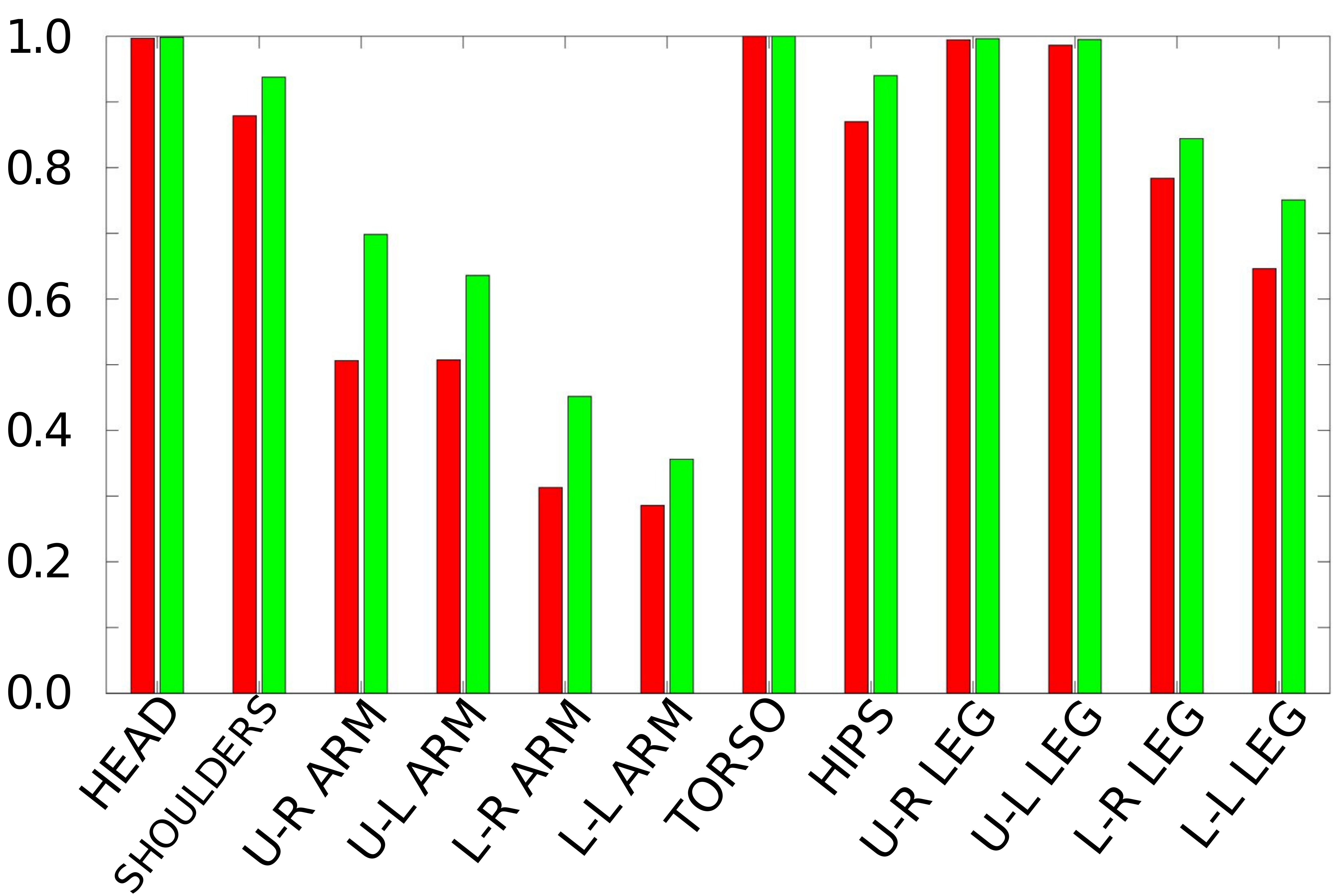}}\
}
\caption{PCP3D scores (\%) for individual parts obtained by FMP\cite{262_yangramanan2013} (red) and ours (green) on both datasets. \textit{(U-L: upper left, U-R: upper right, L-L: lower left, L-R: lower right)}}
\label{fig:partBasedPCP3Dscores}
\end{figure*}

\begin{figure*}[t] \centering
    \includegraphics[width=15cm]{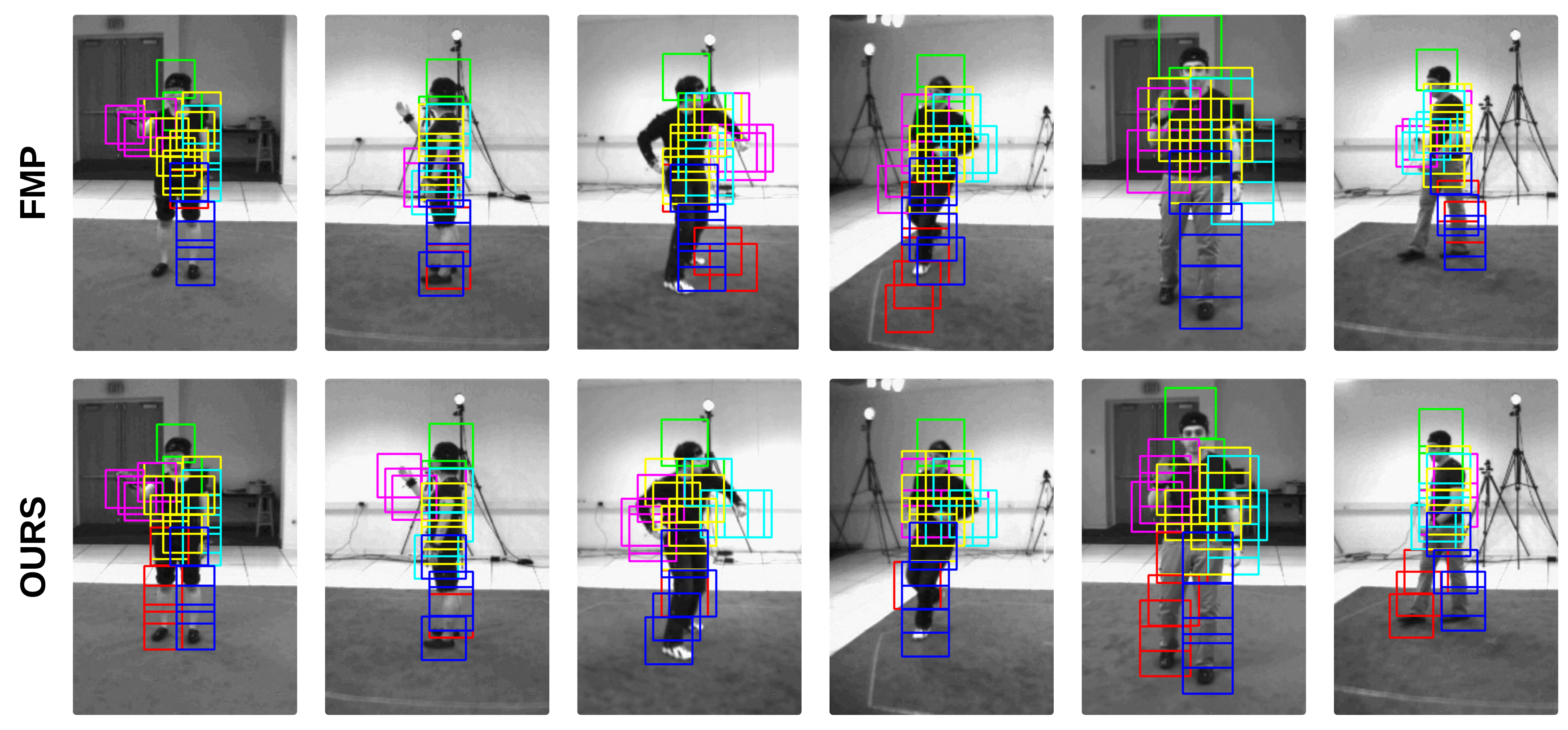}
    \caption{Qualitative comparison of all three subjects performing various activities from different viewpoints. Top: poses obtained with the single-view model \cite{262_yangramanan2013}. Bottom: poses obtained with multi-view pose estimation. Best viewed in colour. 
    \label{qualeval}
    }
\end{figure*}

\begin{figure}[tb]
\centering
\subfloat[HumanEva, S1, all sequences]{\label{fig:pcp3dcurve_he}\includegraphics[width=.475\columnwidth]{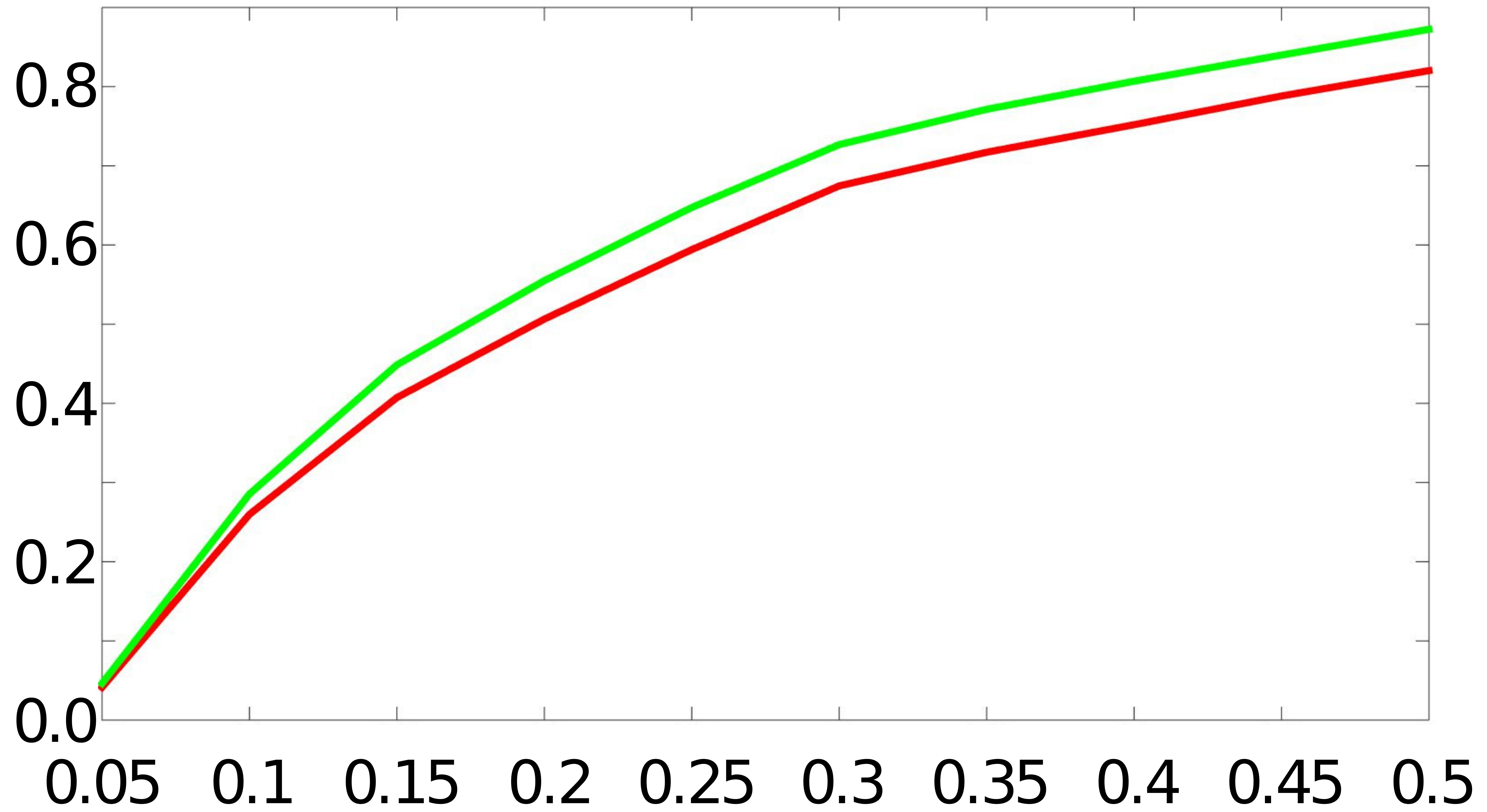}}
\subfloat[UMPM, all sequences]{\label{fig:pcp3dcurve_umpm}\includegraphics[width=.475\columnwidth]{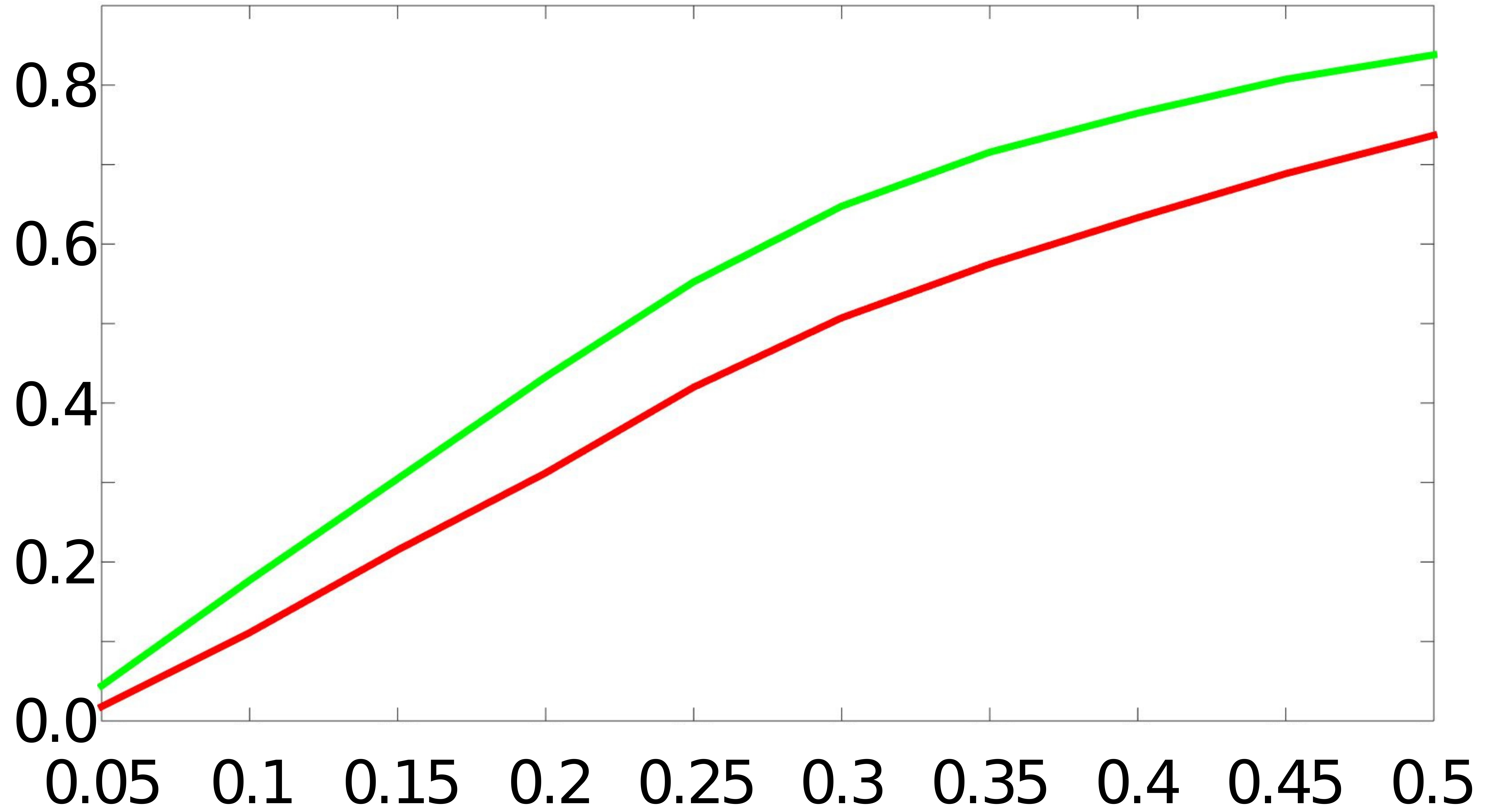}}
\caption{PCP3D curves (\%) obtained by FMP\cite{262_yangramanan2013} (red) and ours (green) on both datasets, as a function of PCP threshold $\gamma$, which controls the ratio of the detected part segments to the ground truth to be considered correctly detected, as defined in Eq. (\ref{eq:pcpThreeD}).}
\label{fig:pcp3dcurves}
\end{figure}

To learn the weights of the error estimating CNN $\sigma_i(\cdot)$, training data sets were augmented with horizontal flip, Gaussian blur and additive noise. As mentioned earlier, we used a finetuned version of VGG-16 \cite{500_SimonyanVGG2014_ICLR} model using pre-trained weights on \emph{ImageNet} to estimate the part-based error of the single-view pose. We removed the fully connected layers and introduced our top model with a hidden layer of 1024 nodes, an output layer of $K$ nodes and parametric ReLU as non-linearity. First, weights of the complete VGG-16 network were frozen so that they are unaffected by the backpropagation and weights of the top model were roughly learnt with a high learning rate. Then, the top model were initialised with these weights, and the last convolutional blocks (namely the last three \emph{conv3-512} layers) were unfrozen for finetuning. We preferred stochastic gradient descent as optimisation algorithm with small learning rate to ensure that the weights of the last convolutional block are marginally updated. To prevent overfit to augmented data sets we applied strong regularisation and also employed \emph{Dropout} \cite{501_SrivastavaDropout2014_JMLR} with a probability of 0.5.

For each multi-view arrangement, i.e. pair combinations of cameras, two pose estimations are produced. Since each view belongs to several multi-view arrangements, we end up with several pose candidates for the same viewpoint, e.g. we obtain two pose candidate for C1, once from the C1-C2 pair and once from the C3-C1 pair. These candidates are simply averaged and obtained 2D poses are triangulated non-linearly to obtain 3D pose for a single time frame. Following the literature on 3D pose estimation \cite{279_burenius2013,301_schick2015} we use the percentage of correctly detected parts in 3D (PCP3D), which is calculated as

\begin{equation} \label{eq:pcpThreeD}
\frac{\lVert \hat{s}_n - s_n \rVert + \lVert \hat{e}_n - e_n \rVert}{2} \leq \gamma \lVert \hat{s}_n - \hat{e}_n \rVert
\end{equation}

\noindent where $s_n$ and $e_n$ are the estimated start and end 3D coordinates of the $n$'th part segment, and $\hat{s}_n$ and $\hat{e}_n$ are the ground truth 3D coordinates for the same part segment. By convention we take $\gamma = 0.5$ in all our computations, unless specified otherwise. 

\textbf{Performances ---} are shown in 
Table \ref{tab:humaneva-yrour} as PCP3D scores on train subject S1 only and over all subjects; while table \ref{tab:umpm-yrour-All} shows PCP3D scores on UMPM test set. We provide three versions: geometric constraints only, geometric and appearance constraints combined, and both constraints with adaptive viewpoint selection. It is clear that in all cases and both data sets, the multi-view scheme significantly improves performance. Depending on the performed action, gains can be significant up to \textbf{9.2\%} in HumanEva and \textbf{10.1\%} in UMPM. The last columns of tables \ref{tab:humaneva-yrour} and \ref{tab:umpm-yrour-All} show that the additional coherence terms decrease the error. Fig. \ref{fig:partBasedPCP3Dscores} demonstrates that this error is distributed over all different parts of the body: we improve most on wrists and elbows, which are important joints for gesture and activity recognition, as seen in table \ref{tab:partBasedPCP3Dscores}. Plots for overall PCP3D curves w.r.t. various thresholds are also given in Fig. \ref{fig:pcp3dcurves}.

\textbf{The adaptive viewpoint selection ---} effectively prevents erroneous consistency terms for certain parts dynamically, due to poor initial single-view estimations as discussed in Section \ref{subsec:wfv}, and shown in table \ref{tab:humaneva-yrour}. 

Fig. \ref{fig_alliterations} depicts intermediate poses and epipolar lines throughout the course of algorithm while Fig. \ref{qualeval} shows several examples from the test set, where faulty poses are corrected with the multi-view approach.  Note that limbs are in particular subject to correction by geometrical and appearance based constraints, since they are considerably susceptible to be mistaken for their respective counterpart. It should be also noted that in case of poor initial detections, a faulty part location can be propagated through the constraints and deteriorate a correct part estimation in other views. Performance tables show that our adaptive viewpoint selection scheme successfully prevents this by considerably decreasing the number of deterioration cases. Particularly, Fig. \ref{fig:improveDegrade} depicts the the amount of improvements and deteriorations w.r.t the baseline, with and without the adaptive viewpoint selection scheme, which efficiently discards the erroneous single-view part detections.

\textbf{Comparison to the state of the art } --- We compare to the original FMP \cite{262_yangramanan2013}, to Schick et al.'s voxel carving based 3D PS method \cite{301_schick2015} and to pre-trained Stacked Hourglass Networks (SHN) \cite{479_newell2016_ECCV}. 

\cite{301_schick2015} report 78\% PCP3D for HumanEva and for all sequences of S1 and S2 (ours: \textbf{83.42\%}) and they report 75\% for UMPM and for all sequences of P1 (ours: \textbf{80.04\%}). 

SHN \cite{479_newell2016_ECCV} requires a cropped input image that is centred around the person with specific scale requirements. Similar to our scheme, 2D poses from different views were triangulated to obtain 3D pose. Table \ref{tab:he_oursVSshn} depicts our estimation performance and two versions of \cite{479_newell2016_ECCV}: First one is with unrestricted images, i.e. same input to our method; and second one with pre-processing steps that require the ground truth. Please note that the SHN heavily depends on the pre-processing, and fails if the person is not centred on image. Our method, which does not require such supervision, obtains similar or better performance to SHN with pre-processed input.

\begin{table}[tb]
\processtable{Comparison of our performance to SHN \cite{479_newell2016_ECCV} on subject 1 of HumanEva dataset, in terms of PCP3D score (\%). First reported result is calculated with unrestricted images (i.e. same input for our method) which is dubbed as \emph{Standard}, while the second one is calculated with cropped input images around the person with a scale requirement, which is dubbed as \emph{Pre-processed}. \textit{(U-L: upper left, U-R: upper right, L-L: lower left, L-R: lower right)} \label{tab:he_oursVSshn}}
{
\vspace{2pt}
\centering
\begin{tabular}{cccc}
\hline \\[-5pt]
Body Part & Ours & SHN Standard & SHN Pre-processed  \\
\hline \\[-5pt]
Head       & 100.0 & 7.40  & 25.17  \\
Shoulders  & 99.47 & 49.08 & 100.0  \\
U-R Arm    & 94.52 & 15.13 & 96.37  \\
U-L Arm    & 88.31 & 45.51 & 98.48  \\
L-R Arm    & 76.09 & 7.73  & 73.98  \\
L-L Arm    & 71.53 & 41.88 & 84.54  \\
Torso      & 100.0 & 52.58 & 100.0  \\
Hips       & 60.63 & 0.0   & 65.52  \\
U-R Leg    & 100.0 & 52.64 & 100.0  \\
U-L Leg    & 100.0 & 51.65 & 100.0  \\
L-R Leg    & 82.69 & 51.65 & 99.41  \\
L-L Leg    & 73.58 & 48.15 & 98.15  \\
\hline \\[-5pt]
Overall    & 87.24 & 35.28 & 86.80  \\
\hline
\end{tabular}}{}
\end{table}

\begin{figure}[tb]
\centering
\subfloat[Without adaptive viewpoint selection]{\label{fig:improveDegrade_Without}\includegraphics[width=\columnwidth]{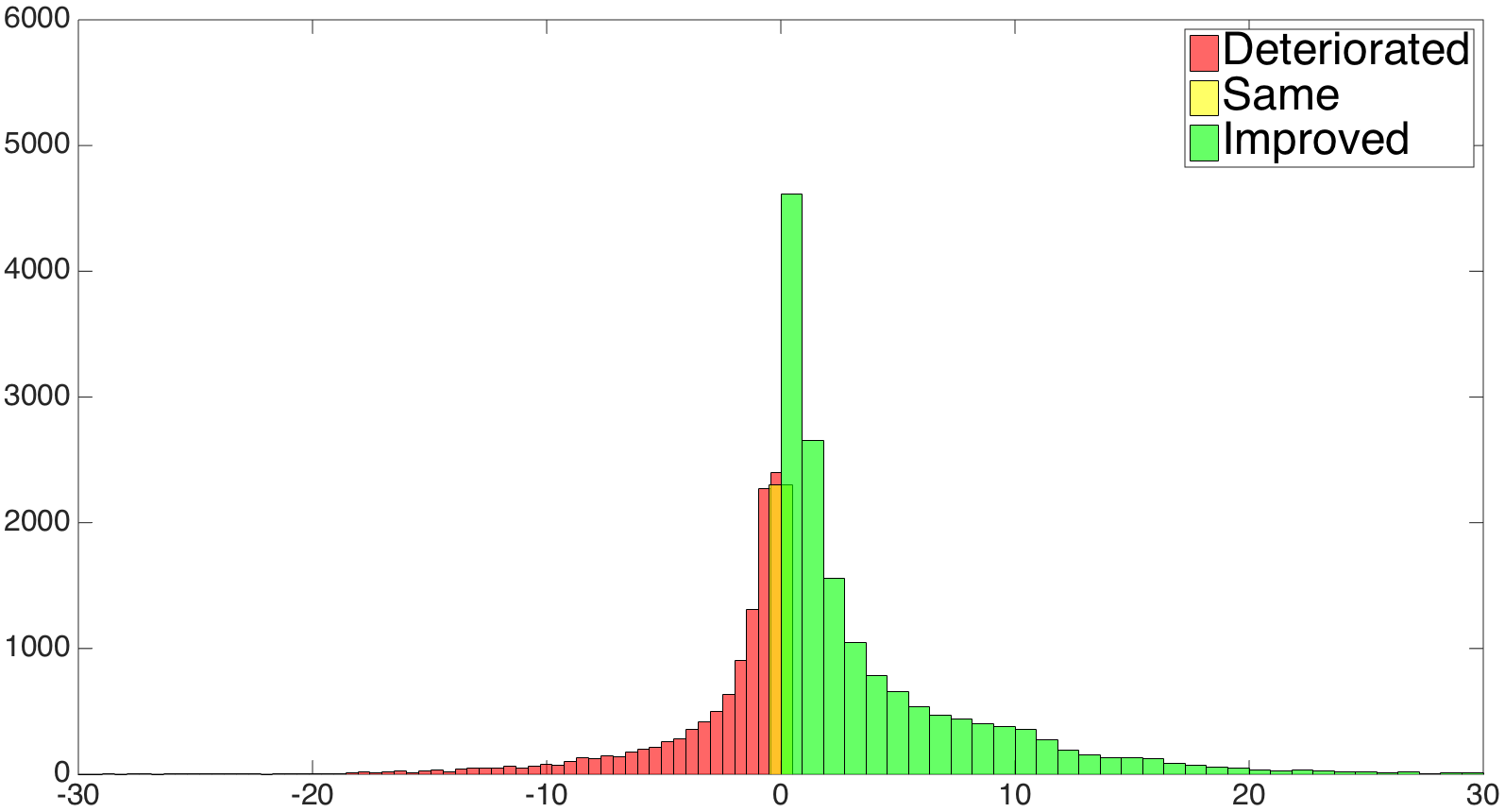}}
\\

\subfloat[With adaptive viewpoint selection]{\label{fig:improveDegrade_With}\includegraphics[width=\columnwidth]{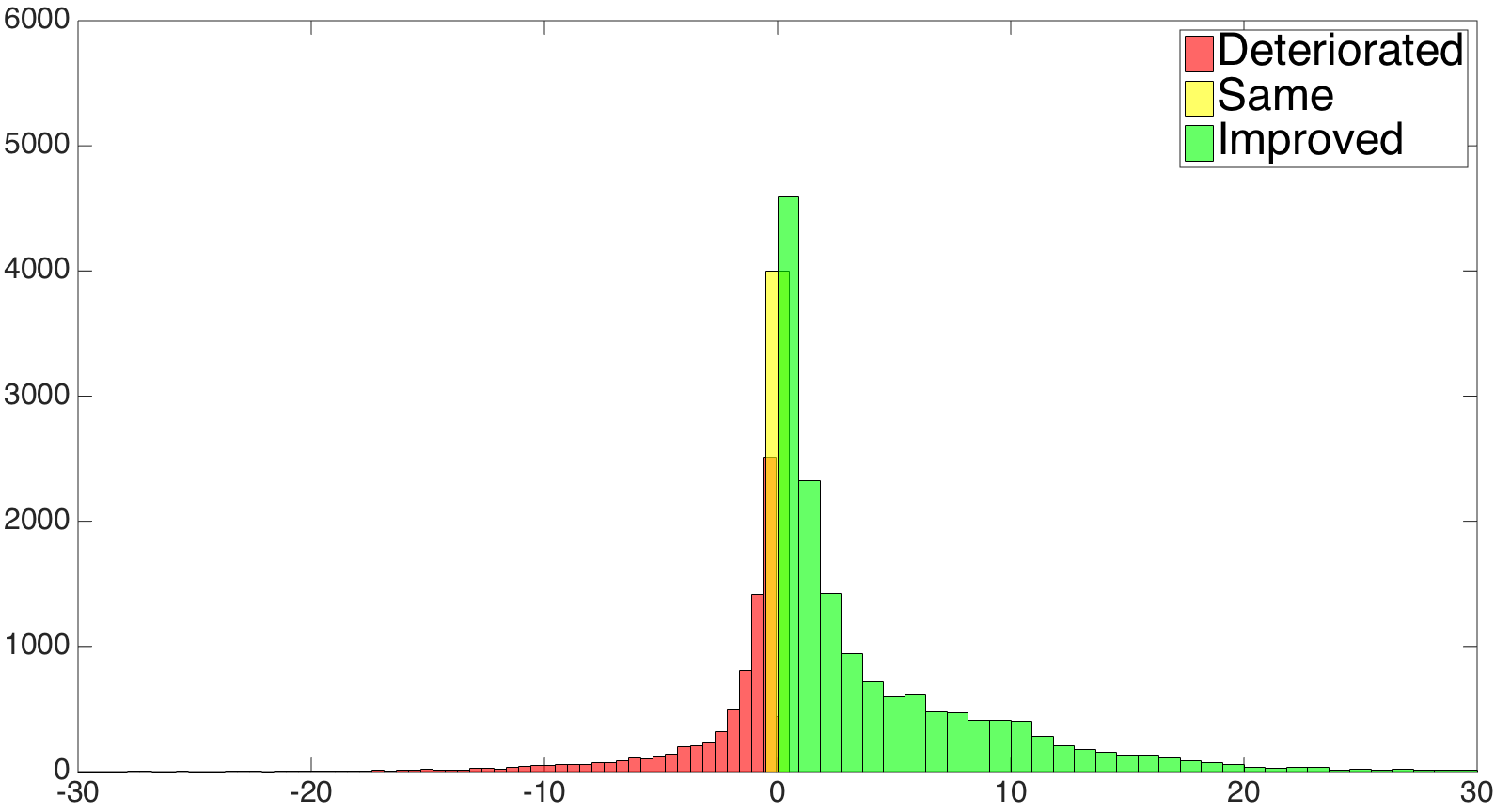}}
\caption{Illustration of the effect of adaptive viewpoint selection. The histograms show differences of errors (px) compared to the baseline \cite{262_yangramanan2013}. Negative differences (red) indicate that our method performs worse, positive differences (green) indicate that our method yielded a better pose. The adaptive mechanism reduces deteriorations while keeping improvements.}
\label{fig:improveDegrade}
\end{figure}

\paragraph{Fast parallel implementation} Our implementation is based on our port of the Matlab/C++code from the single-view method by \cite{262_yangramanan2013} to 100\% pure C++, where crucial parts have also been ported to GPU processing using NVIDIA's CUDA library. This sped up runtime from 3000ms/frame to 880ms/frame on a computer equipped with a 2.4Ghz Xeon E5-2609 processor and an NVIDIA 780 Ti GPU for the single-view algorithm (for a 172x224 image with 32 levels of down-sampling). The multi-view algorithm is slower as 5.73 iterations are performed in average before the results are stable. We are currently working on additional optimisations of computational complexity using approximative parallel implementations of the distance transform on GPUs.

\section{Conclusion}
\label{sec:conclusion}

We proposed a novel multi-view method to estimate articulated body pose from RGB images. Experiments show that combining appearance constraints with geometrical constraints and adaptively applying them on individual parts yields better results than the original single-view model. We also show that our algorithm performs more accurately regardless of the view combinations, and it generalises well in a way to handle unseen subjects and activities.

We plan to extend and evaluate our method in settings with three or more viewpoints, which should be straightforward. Generally, a graph modelling the possible interactions could possibly have high-order cliques, where each clique contains nodes corresponding to the possible views. In practice, it is unsure whether high-order interactions should provide more powerful constraints then (sub)-sets of pairwise constraints. A straightforward algorithm should be similar to the one proposed in the paper: optimizations are carried out over a single-view including pairwise terms involving the different (multiple) support views.

Another improvement would be the extension to a non-calibrated setting, by exploring the self-calibration and epipolar line estimation techniques, which would allow our method to be used in multi-agent robotic systems.

\section{Acknowledgment}
This work was partly supported by a grant of type \emph{BQR} by \emph{Fondation INSA-Lyon}, project \emph{CROME}.


\bibliographystyle{iet}
\bibliography{refs}

\end{document}